\documentclass{article}
\usepackage{spconf,graphicx}

\usepackage{enumitem}
\usepackage{times}
\usepackage{epsfig}

\usepackage{amsmath,amssymb,amsfonts}
\usepackage{booktabs}
\usepackage{multirow}

\usepackage{caption} 

\usepackage{subfig}
\usepackage{tikz}
\usepackage{todonotes}

\usepackage[breaklinks=true,bookmarks=false]{hyperref}
\usepackage[capitalise,nameinlink,noabbrev]{cleveref}
\hypersetup{
 colorlinks=true,
 linkcolor=black,
 urlcolor=cyan,
 citecolor=black
}

\setlist{nosep, leftmargin=14pt}

\usepackage{mwe} 

\title{CLIMAT: Clinically-Inspired Multi-Agent Transformers \\ for Knee Osteoarthritis Trajectory Forecasting}
\name{Huy Hoang Nguyen$^{1}$, Simo Saarakkala$^{1,2}$, Matthew B. Blaschko$^{3}$, Aleksei Tiulpin$^{1,4,5}$}
\address{$^{1}$University of Oulu, Finland, $^{2}$Oulu University Hospital, Finland\\$^{3}$KU Leuven, Belgium, $^{4}$Aalto University, Finland, $^{5}$Ailean Technology Oy, Finland}

\begin{document}

\Crefname{figure}{Suppl. Figure}{Suppl. Figures}
\Crefname{table}{Suppl. Table}{Suppl. Tables}
\Crefname{section}{Suppl. Section}{Suppl. Sections}

%
\maketitle
\begin{abstract}
In medical applications, deep learning methods are built to automate diagnostic tasks. However, a clinically relevant question that practitioners usually face, is how to predict the future trajectory of a disease (prognosis). Current methods for such a problem often require domain knowledge, and are complicated to apply. In this paper, we formulate the prognosis prediction problem as a one-to-many forecasting problem from multimodal data. Inspired by a clinical decision-making process with two agents -- a radiologist and a general practitioner, we model a prognosis prediction problem with two transformer-based components that share information between each other. The first block in this model aims to analyze the imaging data, and the second block leverages the internal representations of the first one as inputs, also fusing them with auxiliary patient data. We show the effectiveness of our method in predicting the development of structural knee osteoarthritis changes over time. Our results show that the proposed method outperforms the state-of-the-art baselines in terms of various performance metrics. In addition, we empirically show that the existence of the multi-agent transformers with depths of $2$ is sufficient to achieve good performances.
Our code is publicly available at \url{https://github.com/MIPT-Oulu/CLIMAT}.
\end{abstract}
\section{Introduction}
\label{sec:intro}

Clinical diagnosis is made by a treating physician or a general practitioner. These specialists are not radiologists and use their services in decision-making. One of the typical problems that such doctors face, is to make an accurate estimation of the disease trajectory (prognosis) based on patient data, findings from imaging, and auxiliary information, such as blood tests. This is an especially relevant task in the case of degenerative disorders. This paper tackles prognosis prediction in knee osteoarthritis (OA) -- the most common musculoskeletal disorder~\cite{glyn2015osteoarthritis}.

\begin{figure}[t!]
    \setlength{\belowcaptionskip}{-12pt}
    \centering
        \subfloat[BL - KL $0$]{\includegraphics[width=0.1\textwidth]{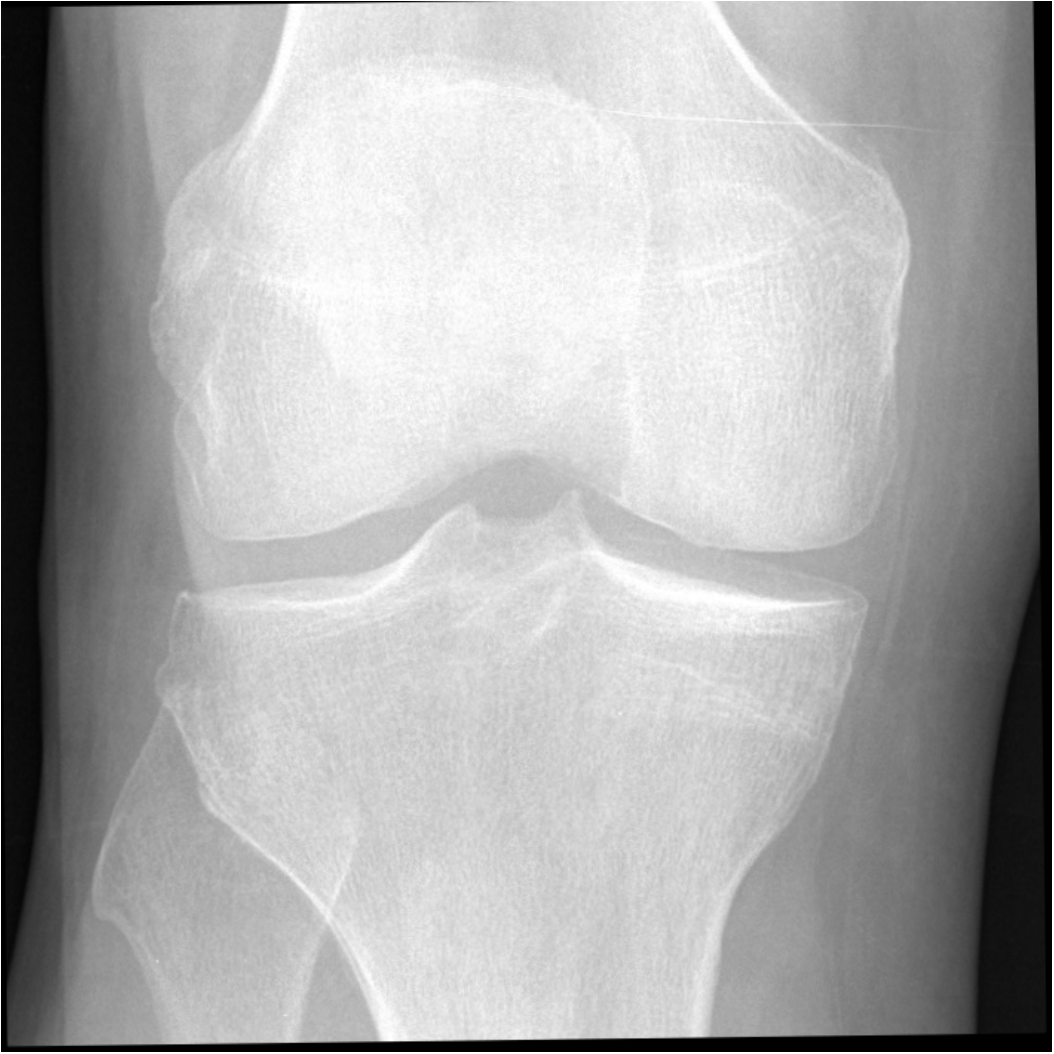}}\hfill%
        \subfloat[Y3 - KL $1$]{\includegraphics[width=0.1\textwidth]{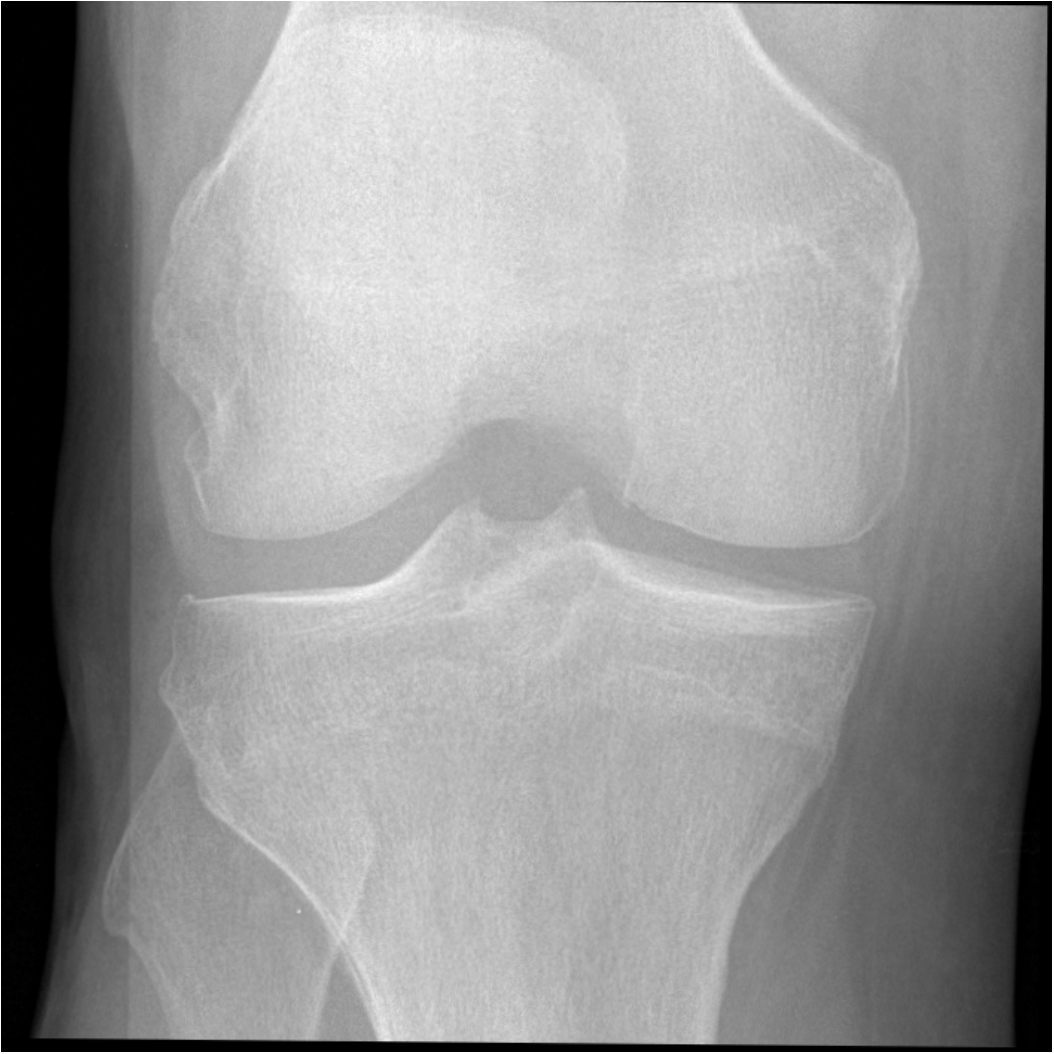}}\hfill%
        \subfloat[Y$6$ - KL $3$]{\includegraphics[width=0.1\textwidth]{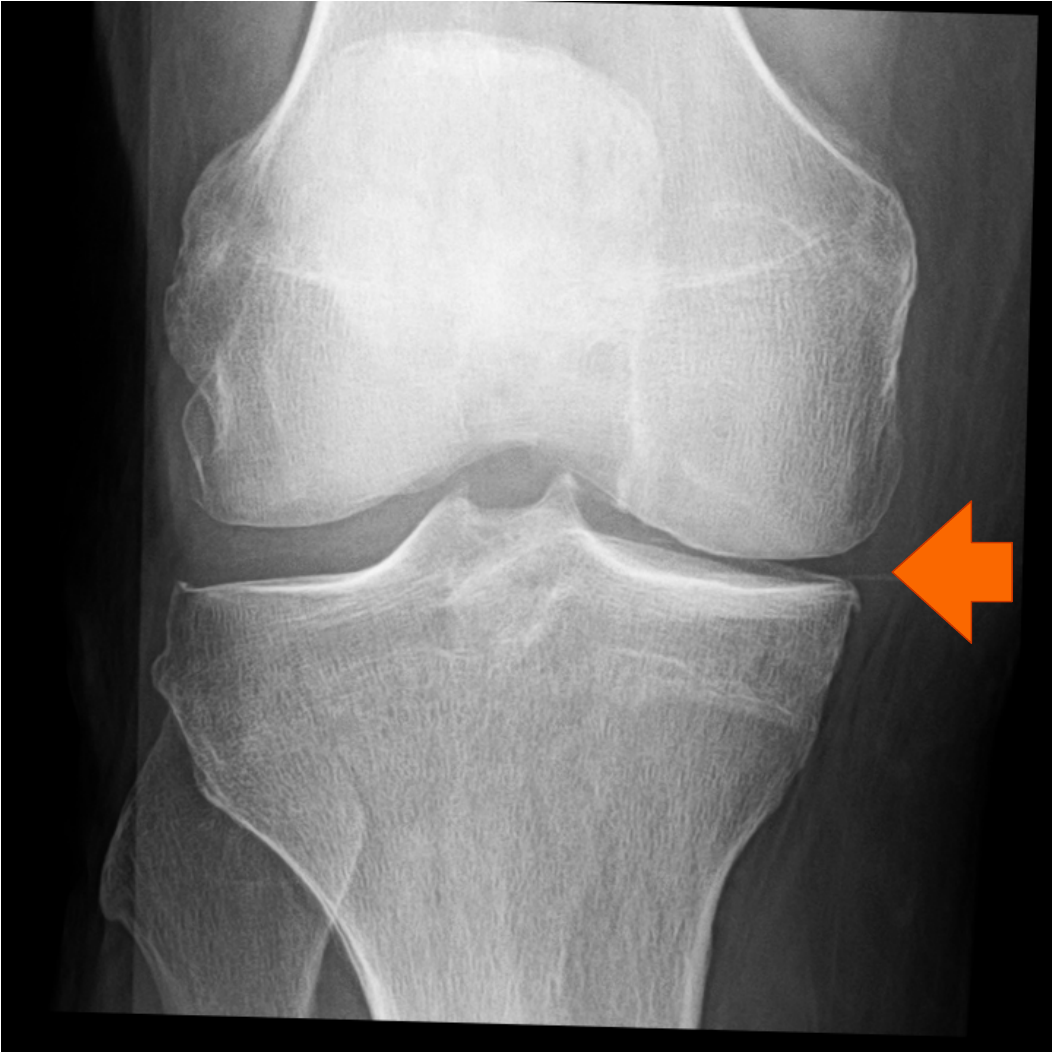}}\hfill%
        \subfloat[Y$8$ - TKR]{\includegraphics[width=0.1\textwidth]{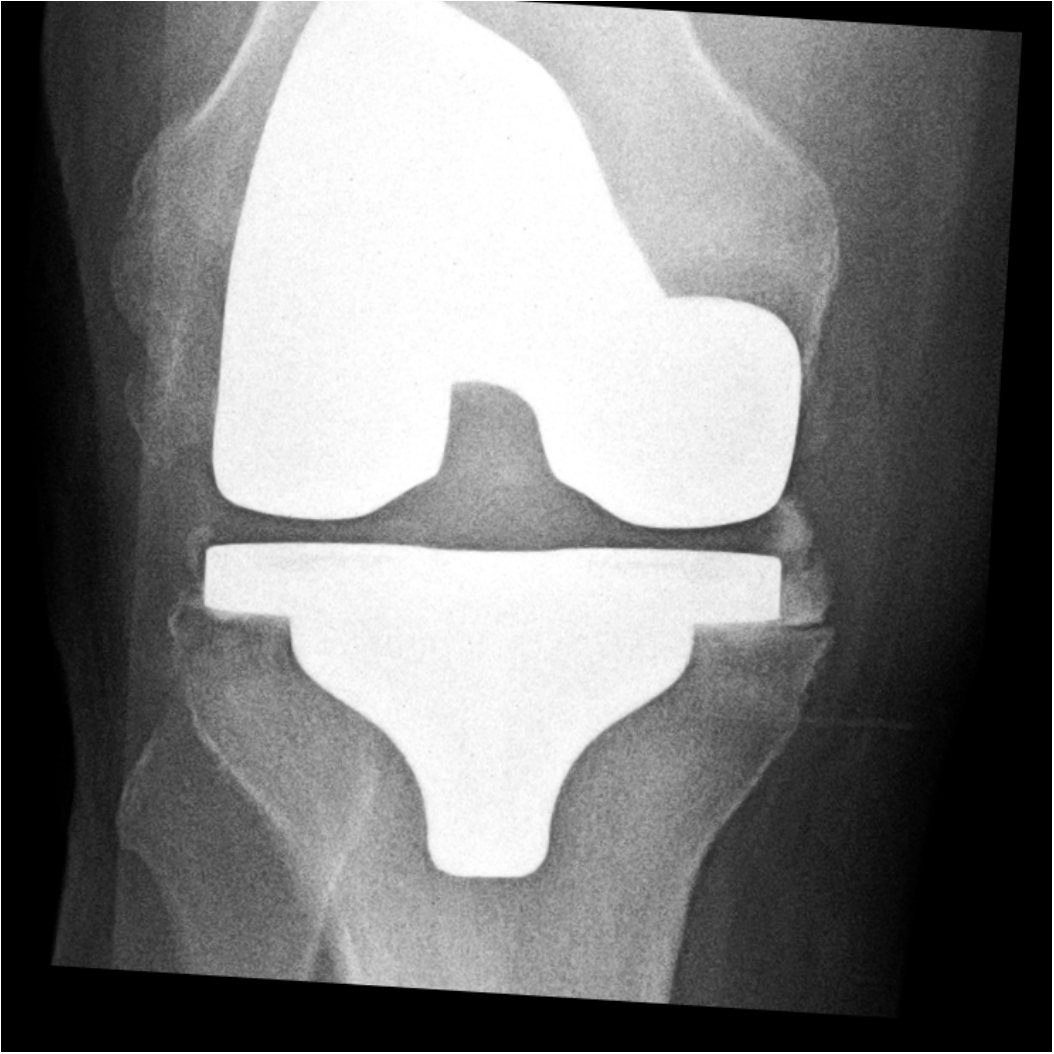}}\hfill%
    \caption{\small Radiographs of a patient with OA progressed in $8$ years. Red arrow indicates joint space narrowing. The disease progressed from Kellgren-Lawrence (KL) grade $0$ at the baseline (BL) to $3$ in $6$ years. At the $8$th year, the patient underwent a total knee replacement (TKR) surgery.}
    \label{fig:progression_sample}
\end{figure}

Among all the joints in the body, OA is mostly prevalent in the knee~\cite{heidari2011knee}. OA is characterized by the breakdown of knee joint cartilage, the appearance of osteophytes, and the narrowing of joint space~\cite{heidari2011knee}, which are imaged using X-ray (radiography). The disease severity is graded according to the Kellgren-Lawrence system~\cite{kellgren1957radiological} from 0 (no OA) to 4 (end stage OA) as shown in~\Cref{fig:samples_by_kl}. Unfortunately, OA progresses over time (depicted in~\cref{fig:progression_sample}) and no cure has yet been developed for OA. However, prediction of disease evolution at an early stage may enable slowing it down, for example using behavioral interventions~\cite{tiulpin2019multimodal}. 


Literature shows that there is a lack of studies on prognosis prediction. From an ML perspective, a more conventional setup is to predict \textit{whether} the patient has the disease~\cite{tiulpin2019multimodal,widera2020multi,guan2020deep}. However, prognosis prediction aims to answer \textit{whether} and \textit{how} the disease would evolve over time. Furthermore, in a real life situation, the treating physician makes the prognosis while interacting with a radiologist or other stakeholders who can provide information (e.g.\ blood tests or radiology reports) about the patient's condition~\cite{jans2013optimizing}. We believe that informing prediction model design with this prior knowledge is valuable, and may provide performance benefits.

In this paper, we propose a Clinically-Inspired Multi-Agent Transformers (CLIMAT) framework, which aims to mimic the interaction process between a general practitioner or treating physician and a radiologist. The \emph{core novel idea} leading to our model design is that a radiologist first analyzes the the image, and provides a radiology report to the doctor who makes the prognosis, also taking into account additional data. In our system, a radiologist module, consisting of a feature extractor (convolutional neural network; CNN) and a transformer, analyses the input imaging data and extracts feature vectors per every image superpixel. Subsequently, the set of superpixels with positional encodings is passed to a transformer that aims to predict the current disease severity stage, characterized by imaging findings. The states of this transformer are fused with auxiliary patient clinical data, and passed to a general practitioner-corresponding transformer module, predicting the disease's severity trajectory. To summarize, our contributions are the following:
\begin{enumerate}
    \item  We propose CLIMAT, a clinically-inspired transformer-based framework that can learn to forecast disease severity from multimodal data in an end-to-end manner. 
    \item From a clinical perspective, to our knowledge, we show the first study on predicting a fine-grained prognosis of knee OA directly from raw imaging data and clinical variables.
    \item We empirically demonstrate superior performance of our method compared to the state-of-the-art baselines.  
\end{enumerate}

\begin{figure}[t!]
    \setlength{\belowcaptionskip}{-10pt}
    \IfFileExists{figures/DeepProg/CLIMAT_workflow-crop.pdf}{}{\immediate\write18{pdfcrop 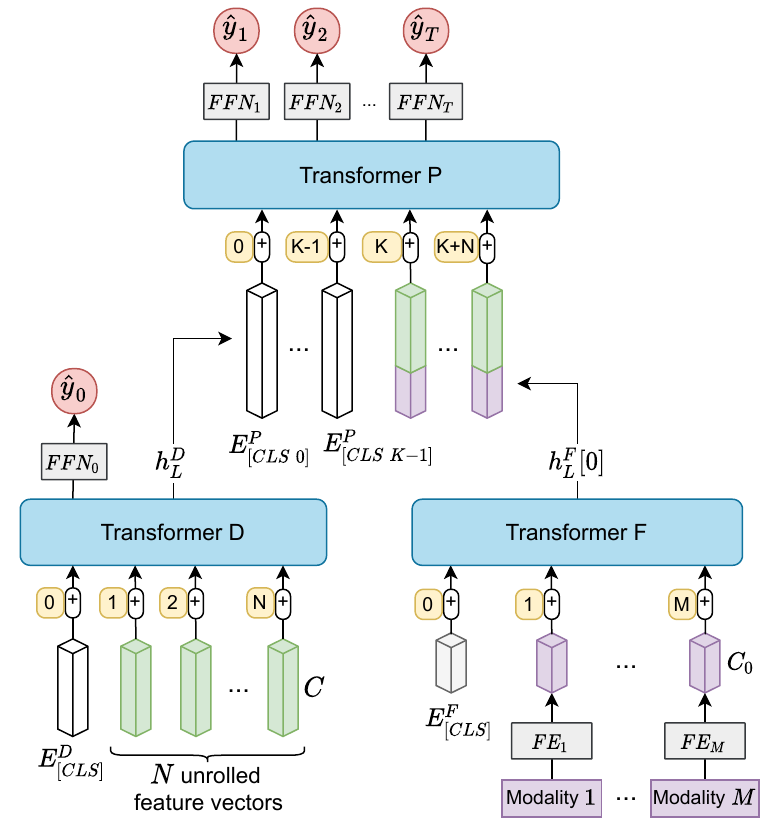}}
        %
         \hspace*{\fill}
        \includegraphics[width=0.38\textwidth]{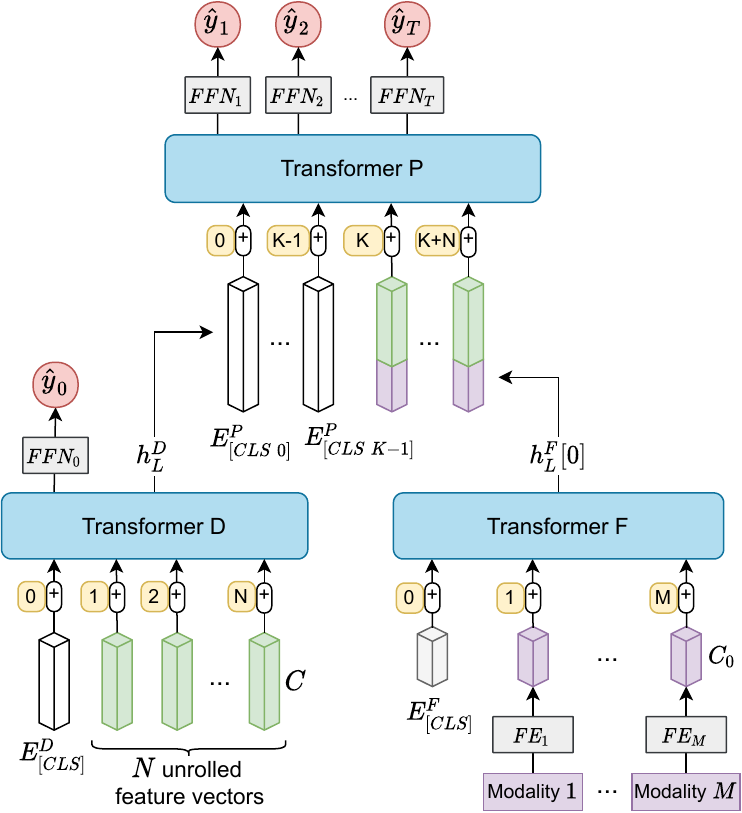}
        \hspace*{\fill}
    \caption{\small The architecture of CLIMAT consists of three transformers. The transformer D as a radiologist performs a diagnosis for the current stage $\hat y_0$ of a disease from visual features. The combination of the transformers F and P, mimicking a general practitioner, aims to forecast future stages $\hat y_{1:T}$ of the disease based on the output states $v_0$ of the transformer D and auxiliary data.}
    \label{fig:deepprog}
\end{figure}

\section{Method}
\subsection{Overview}

\label{sc:conceptual_model}
We model multi-agent decision-making as follows. A radiologist analyzes a medical image (e.g.\ a radiograph) of a patient to provide an interpretation with rich visual description and annotations, allowing the diagnosis of the current stage of the disease. Subsequently, the general practitioner relies on the clinical data (e.g.\ questionnaires or symptomatic assessments), and the provided radiologic interpretation to make a further interpretation if needed to predict the course of the disease in the future. 

In~\cref{fig:deepprog}, we present the workflow of CLIMAT comprising three transformers -- namely D, F, and P, which are the abbreviations for Diagnosis, Fusion, and Prognosis respectively. Specifically, the transformer D acts as the radiologist to perform visual reasoning from imaging data and predict the current stage $\hat y_0$ of the knee OA disease. The other two transformers are responsible for data fusion and forecasting. As such, the transformer F aims to extract a context embedding from clinical variables. Subsequently, the transformer P utilizes the combination of the context embedding and the output states of the transformer D to forecast the disease trajectory $\hat y_{1:T}$. 

\subsection{Multi-output-head transformer}
A transformer encoder comprises a stack of $L$ multi-head self-attention layers, whose input is a sequence of vectors $\{s\}_{i=1}^{N}$ where $s_i \in \mathbb{R}^{1\times C}$, and $C$ is the feature size. We define a transformer with regard to the number of output heads. As such, a transformer with $K$ output heads ($K \ge 1$) is formulated as
\begin{align}
    h_0 & = [E_{[CLS\ 0]},\dots,E_{[CLS\ K-1]}, s_1, \dots, s_N] + E_{[POS]}, \label{eq:h_0}\\
    z_{l-1} & = \textsc{MSA}(\textsc{LN}(h_{l-1})) + h_{l-1}, \\
    h_{l} & = \textsc{MLP}(\textsc{LN}(z_{l-1})) + z_{l-1}, \quad l=\{1, \dots, L\}
\end{align}
where $E_{[CLS\ k]} \in \mathbb{R}^{1 \times C}$ is a learnable token with $k=0\dots K-1$, and $E_{[POS]} \in \mathbb{R}^{(N+K) \times C}$ is a learnable positional embedding. $\textsc{MLP}$ is a multi-layer perceptron (i.e.\ a fully-connected network), $\textsc{LN}$ is a layer normalization~\cite{ba2016layer}, and $\textsc{MSA}(\cdot)$ is a multi-head self-attention layer~\cite{vaswani2017attention}. We take the first $T$ representations in the last layer to perform multi-task predictions via non-linear layers. In general, $K$ is chosen such that $T\leq K+N$. We typically set $K$ to $1$ or $T$. 

\subsection{CLIMAT for knee OA trajectory prediction}

We firstly extract $H\times W \times C$ visual representations of the input radiograph using a stack of convolutional blocks, and then reshape them to $N \times C$, where $N=HW$. Having in mind the idea of visual reasoning, we treat the reshaped representations as an $N$-length sequence of $1\times C$ vectors, and pass them through a transformer, which predicts $y_0$. 
As we convert the image classification problem into a sequence classification one, we include two common ingredients: a sequence start vector, denoted by $E^D_{[CLS]}$, and also the positional embeddings for every super-pixel~\cite{vaswani2017attention,dosovitskiy2020image}. Both of these are learnable vectors, and positional embeddings are added to the superpixels of an image. Once the input sequence of superpixels is passed through the transformer D, we take its first element and pass it through a fully connected network, similar to~\cite{dosovitskiy2020image}.


As the module for predicting the prognosis can utilize other auxiliary modalities, we acquire another transformer -- named F -- to fuse them. Firstly, we project them into the same $C_0$-dimensional feature space using separate feature extractors ($\{FE\}_{m=1}^M$ in~\cref{fig:deepprog}), consisting of a fully connected layer, a ReLU activation, and a layer normalization~\cite{ba2016layer}. Similar to the transformer D, we include an initial embedding $E^F_{[CLS]}$ and a positional embedding to derive the input for the transformer F. Finally, we select the first vector $h^F_L[0]$ in the last layer of the transformer F as a context token representing all the modalities.



As soon as the  context token $h^F_L[0]$ of length $C_0$ is acquired from the context network, we concatenate $N+1$ copies of the token into the last states $h^D_L$ of the transformer D. The prognosis transformer (or transformer P) has $K$ embeddings $E^P_{[CLS]}$. Thus, its input sequence has a length of $K+N+1 \ge T$. To predict the prognosis $y_1,\dots, y_T$, we pass the first $T$ elements of the last layer of the transformer P through $T$ distinct feed-forward networks (FFNs), each of which comprises a layer normalization followed by two fully connected layers separated by a GELU activation~\cite{hendrycks2020gaussian}.

\subsection{Multi-task learning with missing targets}
In practice, each patient commonly has missing annotations throughout follow-up visits. We can handle such an impaired condition with ease by introducing an indicator function to mask out missing targets. Formally, we minimize the following loss
\begin{equation}
    \mathcal{L} =\sum_{i \in I} \frac{1}{\sum_{t=0}^{T} \mathbb{I}_t^i} \sum_{t=0}^{T} w_t\mathbb{I}_t^i \ell(f_t(x^i),y_t^i),
    \label{eq:l_pn}
\end{equation}
where $I$ is the set of sample indices, $(x_i, y_i)$ is a labeled sample, $f$ is our model, $w_t$ is the weight of task $t$, $f_t$ is the output at task $t$, $\mathbb{I}_t^i$ is an indicator function of sample $i$ at task $t$, and $\ell$ is a cross-entropy loss.

\section{Experiments}

\subsection{Data}

We conducted experiments on the Osteoarthritis Initiative (OAI), is are publicly available at \url{https://nda.nih.gov/oai/}.  $4,796$ participants from $45$ to $79$ years old participated in the OAI cohort, which consisted of a baseline, and follow-up visits up to $132$ months. In the present study, we used all knee images that (i) were annotated for KL grade, (ii) did not include implants, and (iii) were acquired with large imaging cohorts: the baseline, and the $12$, $24$, $36$, $48$, $72$, and $96$-month follow-ups (presented in~\Cref{tbl:oai_data_stats}). 
We followed~\cite{tiulpin2019multimodal,nguyen2020semixup} to extract two knees regions of interest from each bilateral radiograph and pre-process each of them. Subsequently, each pre-processed knee image was resized to $256\times256$. 
Additionally, we utilized age, sex, body mass index (BMI), history injury, surgery, and total Western Ontario and McMaster Universities Arthritis Index (WOMAC) as clinical variables (\Cref{tbl:oai_input}).

The OAI dataset includes data from five acquisition centers, which allowed us to utilize the one-center-out cross-validation procedure: data from 4 centers were used for training and validation, and data from the left-out one for testing. We trained and evaluated at $6$ major time points: the baseline, and $1$, $2$, $3$, $4$, $6$, and $8$ years in the future. For each training set from a group of $4$ centers, we performed a $5$-fold cross-validation strategy (\Cref{tbl:oai_data_desc}).

\begin{figure}[t!]
    \centering
    \setlength{\belowcaptionskip}{-10pt}
    \IfFileExists{figures/OAI_method_comparison/plot_outofsite_methods_pn-crop.pdf}{}{\immediate\write18{pdfcrop 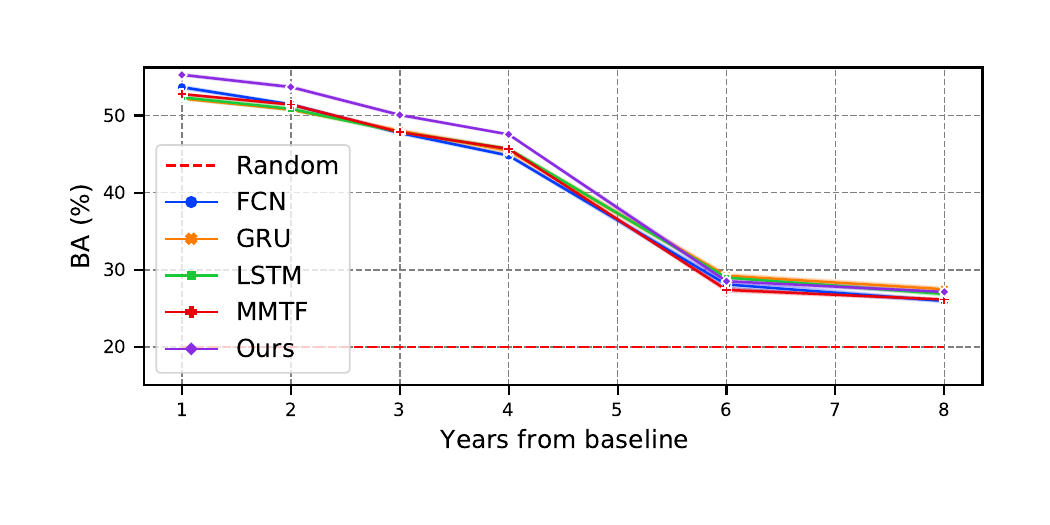}}
    \IfFileExists{figures/OAI_method_comparison/plot_outofsite_methods_pn_rmse-crop.pdf}{}{\immediate\write18{pdfcrop 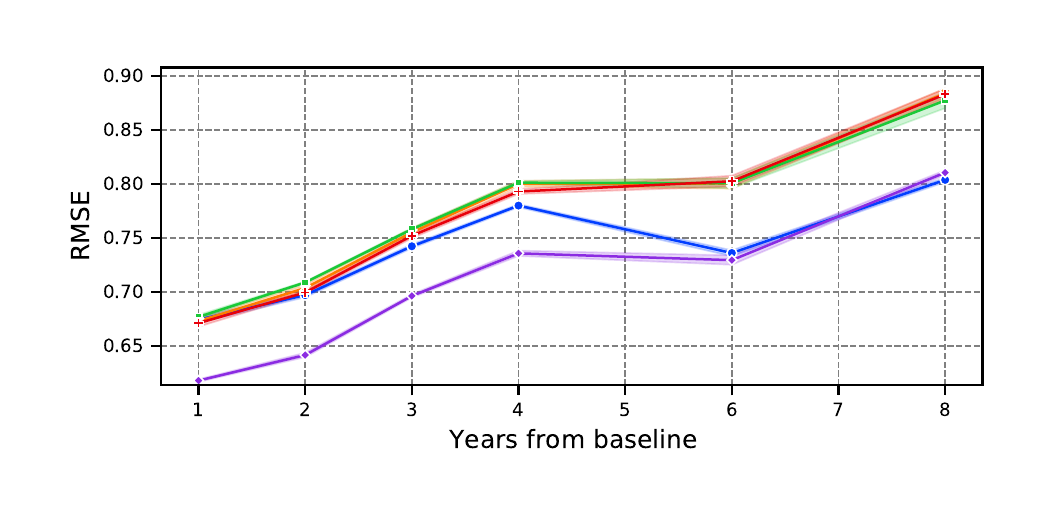}}
    \IfFileExists{figures/OAI_method_comparison/plot_outofsite_methods_pn_ece-crop.pdf}{}{\immediate\write18{pdfcrop 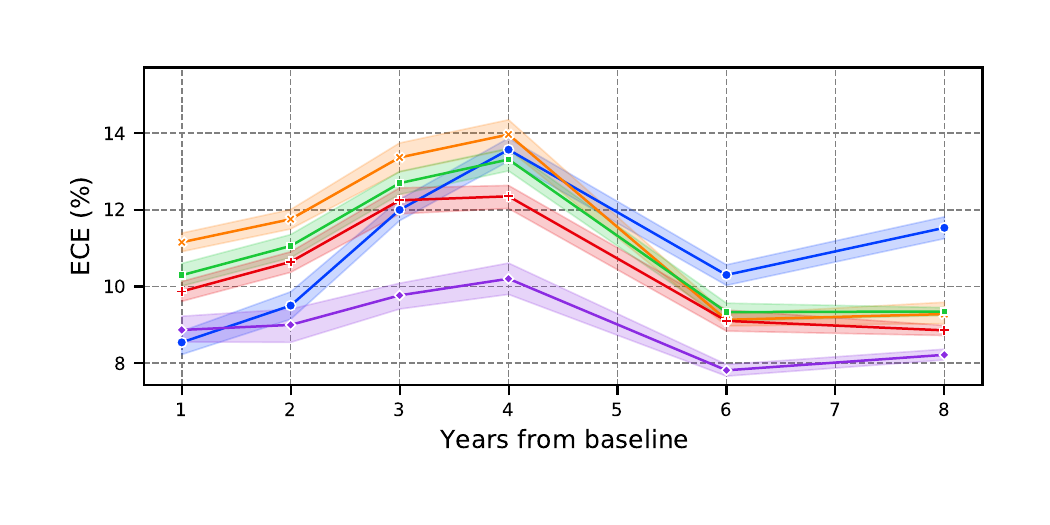}}
    
        \hspace*{\fill}
        \subfloat[]{\includegraphics[width=0.4\textwidth]{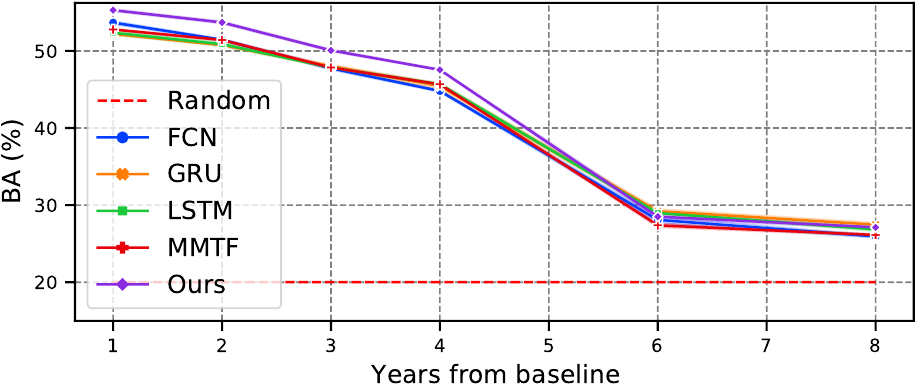}
        \label{fig:oai_ba}
        }
        \hspace*{\fill}
        \\
        \hspace*{\fill}
        \subfloat[]{\includegraphics[width=0.4\textwidth]{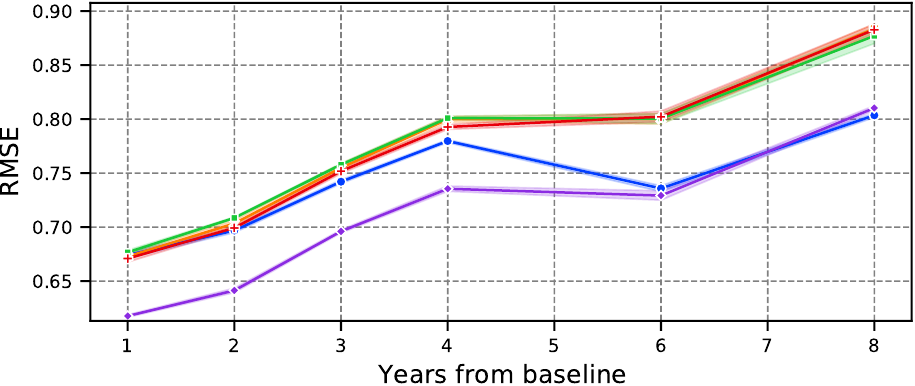}
        \label{fig:oai_rmse}
        }
        \hspace*{\fill}
            
    \caption{\small Performance comparisons on the OAI dataset (average and standard errors over $5$ random seeds).}\label{fig:oai_performance}
        
\end{figure}

\subsection{Experimental Setup}
We trained and evaluated our method and the reference approaches using V100 NVidia GPUs. We implemented all the methods using the Pytorch library, and trained each of them with the same set of configurations. For each problem, we used the Adam optimizer with a learning rate of $1e\mathrm{-}4$ and without any weight decay. The list of augmentations is presented in~\Cref{tbl:train_augmentation}.

To extract visual representations of 2D images, we utilized the convolutional blocks of the ResNet$18$ network~\cite{he2016deep} pretrained on the ImageNet dataset. We used only $1$ [CLS] learnable token ($K=1$) in the transformer P. We used batch size of $128$ for the knee OA experiments. For each scalar numerical or categorical input, we used a common feature extraction architecture with a linear layer, a ReLU activation, and the layer normalization~\cite{ba2016layer}.

Our baselines were models that had the same feature extraction modules for multimodal data, but utilized different architectures to perform discrete time series forecasting. As such, we compared our method to baselines with the forecasting module using fully-connected network (FCN), GRU~\cite{cho2014properties}, LSTM~\cite{hochreiter1997long}, or multimodal transformer (MMTF)~\cite{hu2021transformer}. 
While FCN, MMTF, and CLIMAT are parallel models, GRU and LSTM are sequential ones. Although MMTF and CLIMAT are both based on the self-attention mechanism, our model has a modular structure rather than the flat one as in MMTF. Hyperparameters of the methods are presented in~\Cref{tbl:oai_hyper_params}.


\subsection{Results}


There were significant differences between the performance for predicting the near-future targets (within $4$ years), and also the further ones from the baseline. Results in~\cref{fig:oai_performance} show that our method performed substantially better than the baseline at the near future targets ($t\leq4$). Specifically, in~\Cref{tbl:oai_performance}, our method achieved BAs of $55.3\pm0.2$, $53.7\pm0.2$, $50.1\pm0.2$, and $47.5\pm0.1$ compared to the second-best performances of $53.7\pm0.2$, $51.5\pm0.1$, $48.0\pm0.2$, and $45.7\pm0.2$, respectively. At the far future targets ($t\ge6$), CLIMAT reached RMSEs of $0.73\pm0.005$ and $0.81\pm0.002$ at the $6$ and $8$-year marks respectively, which outperformed the two sequential models and the transformer-based baseline.


\subsection{Ablation studies} 
Since CLIMAT comprises several components, we conducted ablation studies to find out their contributions. Specifically, we investigated how the inclusion of the modular structure of transformers and performing diagnosis prediction help to improve the performance of CLIMAT. Here, we created a baseline model with common feature extraction modules, followed by an LSTM as the sequential reasoning module. We chose LSTM because it is one of the most well-known methods for sequential forecasting problems. 
Subsequently, we replaced LSTM~\cite{hochreiter1997long} by a transformer~\cite{vaswani2017attention}. Then, we used our modular multi-agent transformers instead of the previously flat structure, but the transformer D did not learn from labels of $y_0$. Finally, we utilized the full version of CLIMAT.

\begin{table}[t!]
\centering
\caption{\small Ablation studies.}\label{tbl:ablation_studies}
\scalebox{0.8}{
\begin{tabular}{lcc}
\toprule
\textbf{Feature extractors followed by} & \textbf{Average BA}\\ 
\midrule \midrule
Sequential model & 41.96 \\ 
One flat transformer & 42.01 \\ 
Modular transformers w/o diagnosis & 43.19 \\ 
CLIMAT (Ours) & \textbf{43.47} \\ 
\bottomrule
\end{tabular}}
\vspace{-3mm}
\end{table}

\cref{tbl:ablation_studies} shows that using at least 1 transformer instead of the traditional sequential model improves the performance. Furthermore, the modular multi-agent transformers further helped to increase the performance substantially, especially when we included $y_0$'s in learning. 


From the point of view of transformer depth, we found that increasing it $2$ and $4$ times worsens the performance of prognosis at the early years ($t\le4$), but improves it at later years ($t\ge6$). However, on average, we obtained BAs of $43.47$, $43.43$, and $43.37$ for the depth of $2$, $4$, and $8$, respectively. Thus, the shallow version of the transformer P with only $2$ layers achieved the highest BA. Together with~\cref{tbl:ablation_studies}, it indicates that the existence of the modular structure of transformers matters more than the depth of the transformer P.
\subsection{Interpretability}

\begin{figure}[t!]
    \centering
    \setlength{\belowcaptionskip}{-10pt}
    \IfFileExists{figures/Attention_map/attn_0_59_knee-crop.pdf}{}{\immediate\write18{pdfcrop 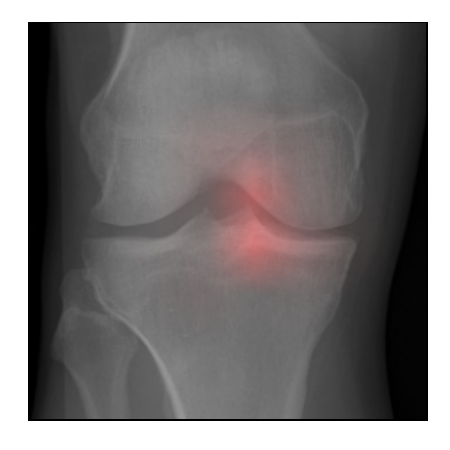}}
    \IfFileExists{figures/Attention_map/attn_0_53_meta-crop.pdf}{}{\immediate\write18{pdfcrop 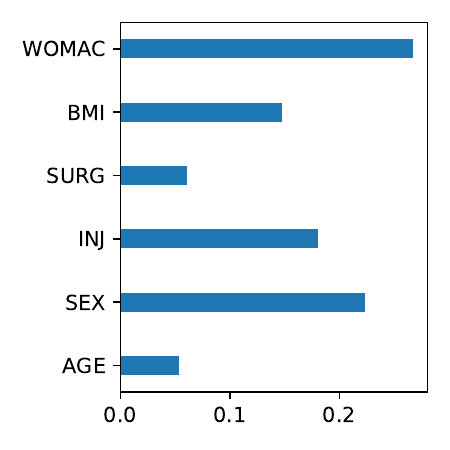}}
    \IfFileExists{figures/Attention_map/attn_0_53_preds-crop.pdf}{}{\immediate\write18{pdfcrop 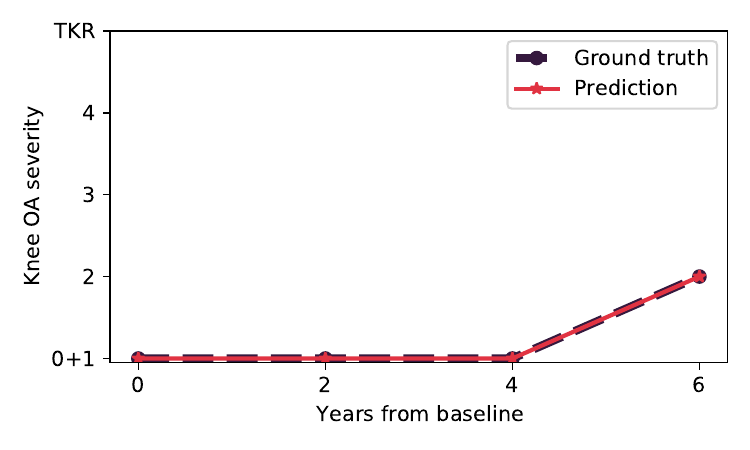}}
        \hspace*{\fill}
        \subfloat[A knee with attenttion maps]{
        \includegraphics[width=0.15\textwidth]{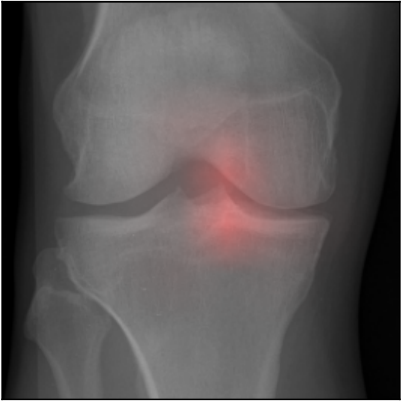}
        \label{fig:oai_img_interpretability}}
        \hfill
        \subfloat[Contributions of the variables]{
        \includegraphics[width=0.15\textwidth]{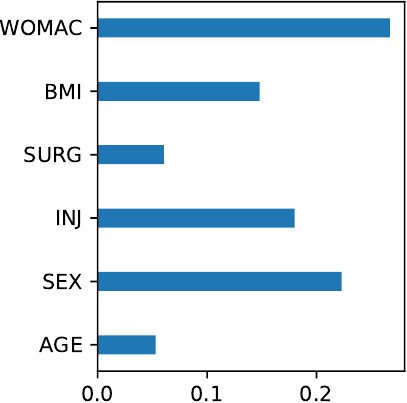}
        \label{fig:oai_meta_interpretability}}  
        \hspace*{\fill}

    \caption{\small An example of progression from a healthy knee at baseline to early osteoarthritis. Our model identified the changes in the intercondylar notch, female sex, and symptomatic status to be the most important factors in predicting progression~\cite{leon2005intercondylar}.}
    \label{fig:oai_interpretability}
\end{figure}
Leveraging the modular structure of CLIMAT, we were able to generate groups of attention maps for different input categories. In~\cref{fig:oai_interpretability}, we present examples of attention maps over a healthy knee X-ray image and the corresponding clinical variables that were extracted from the transformer P and the transformer F, respectively.~\cref{fig:oai_img_interpretability} shows the averages of $4$ self-attention maps. Here, we observe that the final prediction was made based on the changes in the intercondylar notch~\cite{leon2005intercondylar}, as well as the symptomatic evaluation of the patient. \cref{fig:oai_meta_interpretability} indicates that WOMAC, sex, and history of injury were the top-3 impactful clinical variables according to the transformer F for the future knee OA progression of that particular patient. We present more prediction samples in~\Cref{fig:oai_interpretability_more}.

\section{Conclusions}
\label{sc:dis_conc}
In this paper, we proposed a novel transformer-based method to forecast a trajectory of a disease's stage from multimodel data. We applied our method to knee osteoarthritis prognosis prediction problem, and to our knowledge, this is the first study in the realm of OA that tackled this problem. The developed method can be of interest to other fields, where a forecasting of disease course is of interest. 
The main limitation of this study is that our experiments were conducted on a dataset conducted in the research setting. Evaluation of performance of the method in real clinical setting is still needed to understand the value of the method on the OA treatment process. The source code, allowing to fully replicate our results, is publicly available at \url{https://github.com/MIPT-Oulu/CLIMAT}.

\section{COMPLIANCE WITH ETHICAL STANDARDS}
This research study was conducted retrospectively using human subject data made available in open access by Osteoarthritis Initiative (\url{https://nda.nih.gov/oai}). A new ethical approval was not required, as the ethical approval and informed consent of the patients were obtained by OAI and published under open access permission group. 

\section*{Acknowledgments}
The OAI is a public-private partnership comprised of five contracts (N01- AR-2-2258; N01-AR-2-2259; N01-AR-2- 2260; N01-AR-2-2261; N01-AR-2-2262) funded by the National Institutes of Health, a branch of the Department of Health and Human Services, and conducted by the OAI Study Investigators. Private funding partners include Merck Research Laboratories; Novartis Pharmaceuticals Corporation, GlaxoSmithKline; and Pfizer, Inc. Private sector funding for the OAI is managed by the Foundation for the National Institutes of Health. 

The authors wish to acknowledge CSC – IT Center for Science, Finland, for generous computational resources.

We would like to acknowledge the strategic funding of the University of Oulu,  Sigrid Juselius Foundation, Finland.

Dr. Claudia Lindner is acknowledged for providing BoneFinder. Phuoc Dat Nguyen is acknowledged for discussions about transformer.

\bibliographystyle{IEEEbib}
\bibliography{strings,refs}

\begin{thebibliography}{10}

\bibitem{glyn2015osteoarthritis}
Sion Glyn-Jones, AJR Palmer, R~Agricola, AJ~Price, TL~Vincent, H~Weinans, and
  AJ~Carr,
\newblock ``Osteoarthritis,''
\newblock {\em The Lancet}, vol. 386, no. 9991, pp. 376--387, 2015.

\bibitem{heidari2011knee}
Behzad Heidari,
\newblock ``Knee osteoarthritis prevalence, risk factors, pathogenesis and
  features: Part i,''
\newblock {\em Caspian journal of internal medicine}, vol. 2, no. 2, pp. 205,
  2011.

\bibitem{kellgren1957radiological}
JH~Kellgren and JS~Lawrence,
\newblock ``Radiological assessment of osteo-arthrosis,''
\newblock {\em Annals of the rheumatic diseases}, vol. 16, no. 4, pp. 494,
  1957.

\bibitem{tiulpin2019multimodal}
Aleksei Tiulpin, Stefan Klein, Sita~MA Bierma-Zeinstra, J{\'e}r{\^o}me
  Thevenot, Esa Rahtu, Joyce van Meurs, Edwin~HG Oei, and Simo Saarakkala,
\newblock ``Multimodal machine learning-based knee osteoarthritis progression
  prediction from plain radiographs and clinical data,''
\newblock {\em Scientific reports}, vol. 9, no. 1, pp. 1--11, 2019.

\bibitem{widera2020multi}
Pawe{\l} Widera, Paco~MJ Welsing, Christoph Ladel, John Loughlin, Floris~PFJ
  Lafeber, Florence~Petit Dop, Jonathan Larkin, Harrie Weinans, Ali Mobasheri,
  and Jaume Bacardit,
\newblock ``Multi-classifier prediction of knee osteoarthritis progression from
  incomplete imbalanced longitudinal data,''
\newblock {\em Scientific Reports}, vol. 10, no. 1, pp. 1--15, 2020.

\bibitem{guan2020deep}
B~Guan, F~Liu, A~Haj-Mirzaian, S~Demehri, A~Samsonov, T~Neogi, A~Guermazi, and
  R~Kijowski,
\newblock ``Deep learning risk assessment models for predicting progression of
  radiographic medial joint space loss over a 48-month follow-up period,''
\newblock {\em Osteoarthritis and cartilage}, vol. 28, no. 4, pp. 428--437,
  2020.

\bibitem{jans2013optimizing}
LBO Jans, JML Bosmans, KL~Verstraete, and R~Achten,
\newblock ``Optimizing communication between the radiologist and the general
  practitioner,''
\newblock {\em JBR-BTR}, vol. 96, no. 6, pp. 388--390, 2013.

\bibitem{ba2016layer}
Jimmy~Lei Ba, Jamie~Ryan Kiros, and Geoffrey~E Hinton,
\newblock ``Layer normalization,''
\newblock {\em arXiv preprint arXiv:1607.06450}, 2016.

\bibitem{vaswani2017attention}
Ashish Vaswani, Noam Shazeer, Niki Parmar, Jakob Uszkoreit, Llion Jones,
  Aidan~N Gomez, {\L}ukasz Kaiser, and Illia Polosukhin,
\newblock ``Attention is all you need,''
\newblock in {\em Advances in neural information processing systems}, 2017, pp.
  5998--6008.

\bibitem{dosovitskiy2020image}
Alexey Dosovitskiy, Lucas Beyer, Alexander Kolesnikov, Dirk Weissenborn,
  Xiaohua Zhai, Thomas Unterthiner, Mostafa Dehghani, Matthias Minderer, Georg
  Heigold, Sylvain Gelly, et~al.,
\newblock ``An image is worth 16x16 words: Transformers for image recognition
  at scale,''
\newblock {\em arXiv preprint arXiv:2010.11929}, vol. 1, 2020.

\bibitem{hendrycks2020gaussian}
Dan Hendrycks and Kevin Gimpel,
\newblock ``Gaussian error linear units ({GELUs}),'' 2020,
\newblock arXiv:1606.08415.

\bibitem{nguyen2020semixup}
Huy~Hoang Nguyen, Simo Saarakkala, Matthew Blaschko, and Aleksei Tiulpin,
\newblock ``Semixup: In-and out-of-manifold regularization for deep
  semi-supervised knee osteoarthritis severity grading from plain
  radiographs.,''
\newblock {\em IEEE Transactions on Medical Imaging}, vol. 39, no. 12, pp.
  4346--4356, 2020.

\bibitem{he2016deep}
Kaiming He, Xiangyu Zhang, Shaoqing Ren, and Jian Sun,
\newblock ``Deep residual learning for image recognition,''
\newblock in {\em Proceedings of the IEEE conference on computer vision and
  pattern recognition}, 2016, pp. 770--778.

\bibitem{cho2014properties}
Kyunghyun Cho, Bart Van~Merri{\"e}nboer, Dzmitry Bahdanau, and Yoshua Bengio,
\newblock ``On the properties of neural machine translation: Encoder-decoder
  approaches,''
\newblock {\em arXiv preprint arXiv:1409.1259}, 2014.

\bibitem{hochreiter1997long}
Sepp Hochreiter and J{\"u}rgen Schmidhuber,
\newblock ``Long short-term memory,''
\newblock {\em Neural computation}, vol. 9, no. 8, pp. 1735--1780, 1997.

\bibitem{hu2021transformer}
Shi Hu, Egill Fridgeirsson, Guido van Wingen, and Max Welling,
\newblock ``Transformer-based deep survival analysis,''
\newblock in {\em Survival Prediction-Algorithms, Challenges and Applications}.
  PMLR, 2021, pp. 132--148.

\bibitem{leon2005intercondylar}
Heriberto~Ojeda Le{\'o}n, Carlos E~Rodr{\'\i}guez Blanco, Todd~B Guthrie, and
  Oscar J~Nordelo Mart{\'\i}nez,
\newblock ``Intercondylar notch stenosis in degenerative arthritis of the
  knee,''
\newblock {\em Arthroscopy: The Journal of Arthroscopic \& Related Surgery},
  vol. 21, no. 3, pp. 294--302, 2005.

\end{thebibliography}

\renewcommand{\thepage}{S\arabic{page}} 
\renewcommand{\thesection}{S\arabic{section}}  
\renewcommand{\thetable}{S\arabic{table}}  
\renewcommand{\thefigure}{S\arabic{figure}}

\setcounter{page}{1}
\setcounter{figure}{0}
\setcounter{table}{0}
\setcounter{section}{1}

\clearpage

\section*{Supplementary materials}

\begin{table}[h!]
\centering
\caption{Input variables for forecasting knee OA severity.}
\begin{tabular}{|l|l|l|}
\hline
\textbf{Group} & \textbf{Variable name} & \textbf{Data type} \\ \hline \hline
Raw imaging & Knee X-ray & 2D \\ \hline \hline
Clinical  Variables & Age & Numerical \\ \cline{ 2- 3}
\multicolumn{ 1}{|l|}{} & WOMAC & Numerical \\ \cline{ 2- 3}
\multicolumn{ 1}{|l|}{} & Sex & Categorical \\ \cline{ 2- 3}
\multicolumn{ 1}{|l|}{} & Injury & Categorical \\ \cline{ 2- 3}
\multicolumn{ 1}{|l|}{} & Surgery & Categorical \\ \cline{ 2- 3}
\multicolumn{ 1}{|l|}{} & BMI & Numerical \\ \hline
\end{tabular}
\label{tbl:oai_input}
\end{table}

\begin{table}[h!]
  \centering
  \caption{An ordered list of common transformations. ($\checkmark$) indicates transformations only used in the training phase.}
\scalebox{0.9}{
    \begin{tabular}{lcrc}
    \toprule
    \multicolumn{1}{l}{\textbf{Transformation}} & \textbf{Prob.} & \multicolumn{1}{c}{\textbf{Parameter}} \\
    \midrule \midrule
    Center cropping & 1 & $700 \times 700$ \\
    Resize & 1 & $280 \times 280$ \\
    Gaussian noise ($\checkmark$) & 0.5   & 0.3 \\
    Rotation ($\checkmark$) & 1     & [-10, 10] \\
    Random cropping ($\checkmark$) & 1     & $256\times256$ \\
    Center cropping & 1     & $256\times256$ \\
    Gamma correction ($\checkmark$) & 0.5   & [0.5, 1.5] \\
    Z-score stdardization & 1 &  \\
    \bottomrule
    \end{tabular}}
  \label{tbl:train_augmentation}%
\end{table} 

\begin{table}[h!]
  \centering
  \caption{Common and specific hyperparameters for the methods.}
\scalebox{0.9}{
    \begin{tabular}{llcr}
    \toprule 
    \multicolumn{1}{l}{\textbf{Key}} & \textbf{Value}  \\
    \midrule \midrule
    \multicolumn{2}{l}{\textbf{Common}} \\
    \midrule 
    Raw image feature extractor & ResNet$18$ \\
    Number of Conv2D blocks & 5 \\
    \midrule
    Feature length of scalar input & 128 \\
    MLP hidden unit & 256 \\
    Dropout rate & 0.3 \\
    \midrule \midrule
    \multicolumn{2}{l}{\textbf{CLIMAT}} \\
    \midrule
    Number of [CLS] tokens ($K$) & 1 \\
    Depth of transformer D, F, P & 2 \\
    MSA heads of D, F, P &  4 \\
    \bottomrule
    \end{tabular}}
  \label{tbl:oai_hyper_params}%
\end{table}

\clearpage

\begin{figure*}[h!]
    \centering
        \subfloat[KL $0$ ]{\includegraphics[width=0.19\textwidth]{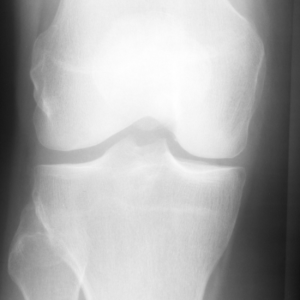}}\hfill%
        \subfloat[KL $1$ ]{\includegraphics[width=0.19\textwidth]{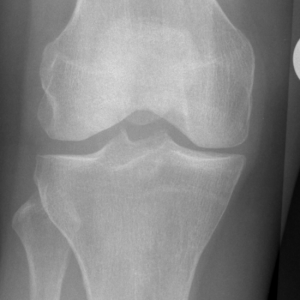}}\hfill%
        \subfloat[KL $2$ ]{\includegraphics[width=0.19\textwidth]{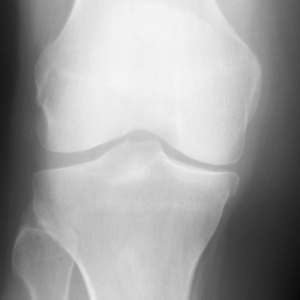}}\hfill%
        \subfloat[KL $3$ ]{\includegraphics[width=0.19\textwidth]{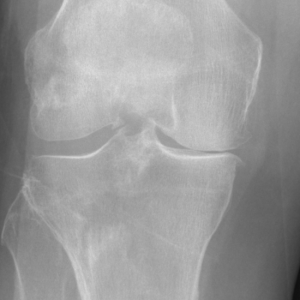}}\hfill%
        \subfloat[KL $4$ ]{\includegraphics[width=0.19\textwidth]{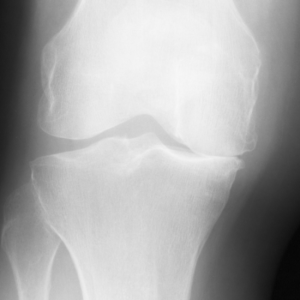}}\hfill%
    \caption{\small The Kellgren-Lawrence (KL) system is commonly used to assess the severity of OA. As such, the system classifies OA into $5$ grades, which correspond to: no sign of OA, doubtful OA, early OA, moderate OA, and severe OA, respectively.}
    \label{fig:samples_by_kl}
\end{figure*}

\begin{table*}[h!]
  \centering
  \caption{Knee OA target statistics over the $6$ primary visits of in the OAI cohort study.}

\addtolength{\tabcolsep}{5pt}   
\begin{tabular}{lrrrrrr}

\toprule
\textbf{Visit} &  \textbf{KL 0} &  \textbf{KL 1} &  \textbf{KL 2} &  \textbf{KL 3} &  \textbf{KL 4} &  \textbf{TKR} \\
\midrule \midrule
Baseline &  3448 &  1597 &  2374 &  1239 &   295 &   61 \\
12 months &  3113 &  1445 &  2221 &  1230 &   355 &   76 \\
24 months &  2893 &  1348 &  2079 &  1172 &   367 &   97 \\
36 months &  2735 &  1252 &  1986 &  1147 &   377 &  135 \\
72 months &  1866 &  1007 &   471 &   201 &    26 &    9 \\
96 months &  1899 &   987 &   488 &   239 &    47 &   15 \\
\bottomrule
\end{tabular}
\label{tbl:oai_data_stats}%
\end{table*}

\begin{table*}[h!]
  \centering
  \caption{Data settings on the OAI dataset across $5$ acquisition sites.}
\begin{tabular}{cccrrrrrrrr}
\toprule
& & & \multicolumn{8}{c}{\textbf{Years from baseline}} \\ \cmidrule{4-11}
     
\textbf{Test site} & \textbf{Phase} & \textbf{Baseline only} &   \textbf{1} &      \textbf{2} &      \textbf{3} &      \textbf{4} &     \textbf{5} &     \textbf{6} &     \textbf{7} &     \textbf{8}    \\
\midrule \midrule
\multirow{2}{*}{A} & \multirow{1}{*}{Training/validation} & Yes &   7155 &   6706 &   6418 &   6173 &     0 &  3077 &     0 &  3147 \\
  & Test & Yes &   1229 &   1195 &   1162 &   1075 &     0 &   497 &     0 &   525 \\
\cmidrule{1-11}
\multirow{2}{*}{B} & \multirow{1}{*}{Training/validation} & Yes &   6572 &   6196 &   5935 &   5671 &     0 &  2797 &     0 &  2851 \\
  & Test & Yes &   1812 &   1705 &   1645 &   1577 &     0 &   777 &     0 &   821 \\
\cmidrule{1-11}
\multirow{2}{*}{C} & \multirow{1}{*}{Training/validation} & Yes &   5902 &   5490 &   5289 &   5023 &     0 &  2442 &     0 &  2537 \\
  & Test & Yes &   2482 &   2411 &   2291 &   2225 &     0 &  1132 &     0 &  1135 \\
\cmidrule{1-11}
\multirow{2}{*}{D} & \multirow{1}{*}{Training/validation} & Yes &   6274 &   5951 &   5706 &   5459 &     0 &  2636 &     0 &  2698 \\
  & Test & Yes &   2110 &   1950 &   1874 &   1789 &     0 &   938 &     0 &   974 \\
\cmidrule{1-11}
\multirow{2}{*}{E} & \multirow{1}{*}{Training/validation} & Yes &   7633 &   7261 &   6972 &   6666 &     0 &  3344 &     0 &  3455 \\
  & Test & Yes &    751 &    640 &    608 &    582 &     0 &   230 &     0 &   217 \\
\bottomrule
\end{tabular}
\label{tbl:oai_data_desc}
\end{table*}

\begin{table*}[htbp]
    \centering
    \caption{Detailed performances on the OAI dataset (average and standard errors over $5$ random seeds).}
    \renewcommand{\arraystretch}{1.2}
    \begin{tabular}{clcc}
\toprule
\textbf{Year} & \textbf{Method} & \textbf{BA (\%) $\uparrow$}  & \textbf{RMSE} $\downarrow$ \\
\midrule \midrule
\multirow{5}{*}{1} & FCN & 53.7$\pm$0.2 & 0.67$\pm$0.003 \\
  & GRU & 52.2$\pm$0.2 & 0.67$\pm$0.003 \\
  & LSTM & 52.4$\pm$0.3 & 0.68$\pm$0.002 \\
  & MMTF & 52.8$\pm$0.1 & 0.67$\pm$0.003 \\
  & Ours & \textbf{55.3$\pm$0.2} & \textbf{0.62$\pm$0.002} \\
\cline{1-4}
\multirow{5}{*}{2} & FCN & 51.5$\pm$0.1 & 0.70$\pm$0.002 \\
  & GRU & 50.8$\pm$0.1 & 0.70$\pm$0.003 \\
  & LSTM & 50.9$\pm$0.3 & 0.71$\pm$0.001 \\
  & MMTF & 51.4$\pm$0.2 & 0.70$\pm$0.002 \\
  & Ours & \textbf{53.7$\pm$0.2} & \textbf{0.64$\pm$0.002} \\
\cline{1-4}
\multirow{5}{*}{3} & FCN & 47.7$\pm$0.2 & 0.74$\pm$0.002 \\
  & GRU & 48.0$\pm$0.2 & 0.76$\pm$0.004 \\
  & LSTM & 47.9$\pm$0.1 & 0.76$\pm$0.001 \\
  & MMTF & 47.8$\pm$0.2 & 0.75$\pm$0.002 \\
  & Ours & \textbf{50.1$\pm$0.2} & \textbf{0.70$\pm$0.002} \\
\cline{1-4}
\multirow{5}{*}{4} & FCN & 44.8$\pm$0.2 & 0.78$\pm$0.002 \\
  & GRU & 45.4$\pm$0.4 & 0.80$\pm$0.003 \\
  & LSTM & 45.7$\pm$0.2 & 0.80$\pm$0.002 \\
  & MMTF & 45.7$\pm$0.0 & 0.79$\pm$0.002 \\
  & Ours & \textbf{47.5$\pm$0.1} & \textbf{0.74$\pm$0.003} \\
\cline{1-4}
\multirow{5}{*}{6} & FCN & 28.1$\pm$0.4 & 0.74$\pm$0.003 \\
  & GRU & \textbf{29.2$\pm$0.3} & 0.80$\pm$0.005 \\
  & LSTM & 29.0$\pm$0.2 & 0.80$\pm$0.005 \\
  & MMTF & 27.4$\pm$0.3 & 0.80$\pm$0.005 \\
  & Ours & 28.5$\pm$0.4 & \textbf{0.73$\pm$0.005} \\
\cline{1-4}
\multirow{5}{*}{8} & FCN & 25.9$\pm$0.2 & \textbf{0.80$\pm$0.002} \\
  & GRU & \textbf{27.5$\pm$0.3} & 0.88$\pm$0.005 \\
  & LSTM & 26.9$\pm$0.3 & 0.88$\pm$0.008 \\
  & MMTF & 26.1$\pm$0.2 & 0.88$\pm$0.006 \\
  & Ours & 27.1$\pm$0.2 & 0.81$\pm$0.002 \\

\bottomrule
\end{tabular}
    \label{tbl:oai_performance}
\end{table*}

\begin{figure*}
    \centering
        \includegraphics[width=0.26\textwidth]{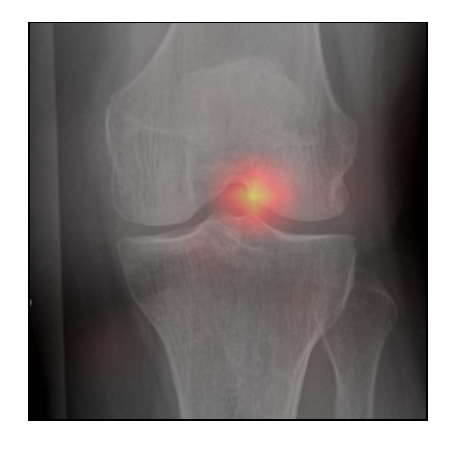}
        \hfill
        \subfloat[]{
        \includegraphics[width=0.26\textwidth]{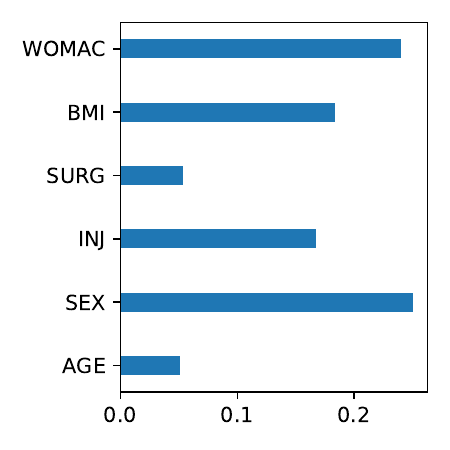}
        } \hfill 
        \includegraphics[width=0.43\textwidth]{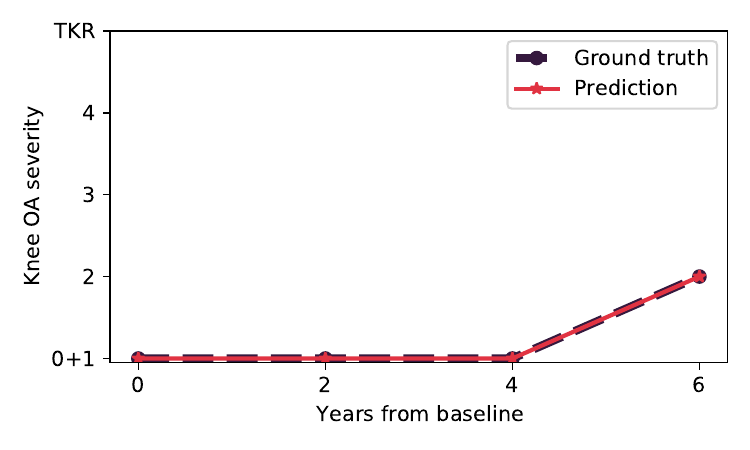}
        \\
        \includegraphics[width=0.26\textwidth]{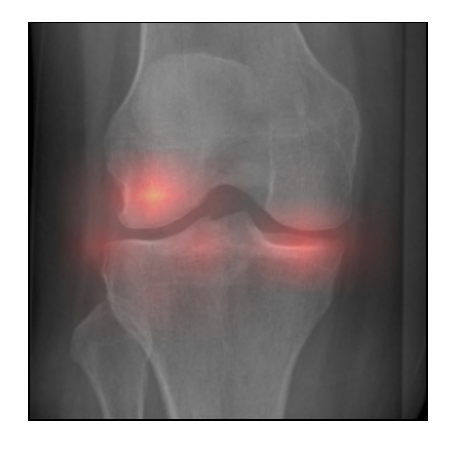}
        \hfill
        \subfloat[]{
        \includegraphics[width=0.26\textwidth]{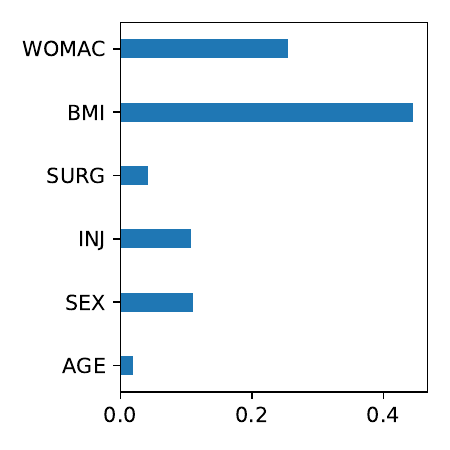}
        } 
        \hfill 
        \includegraphics[width=0.43\textwidth]{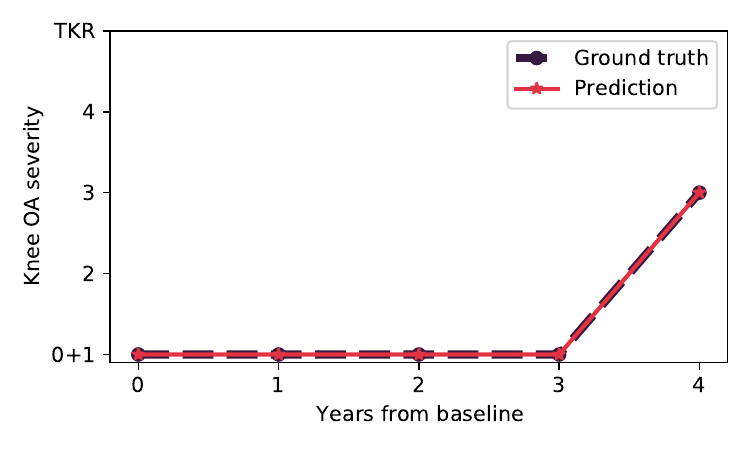}
        \\
        \includegraphics[width=0.26\textwidth]{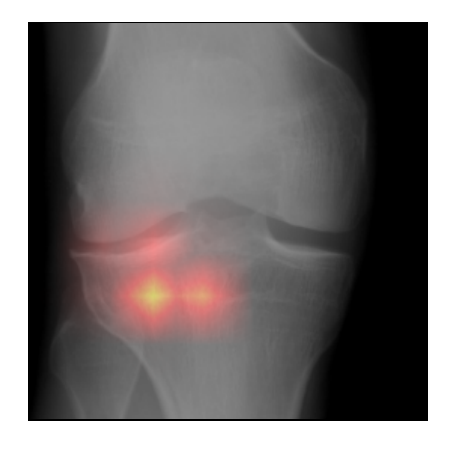}
        \hfill
        \subfloat[]{
        \includegraphics[width=0.26\textwidth]{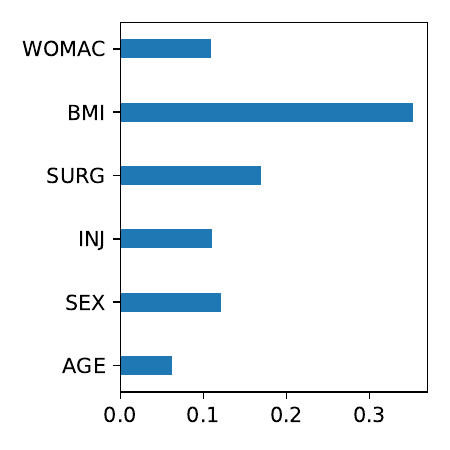}
        } \hfill 
        \includegraphics[width=0.43\textwidth]{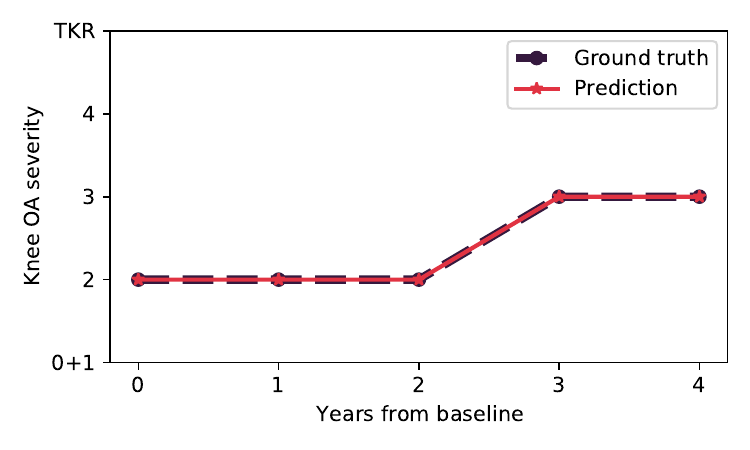}
        \\
        \includegraphics[width=0.26\textwidth]{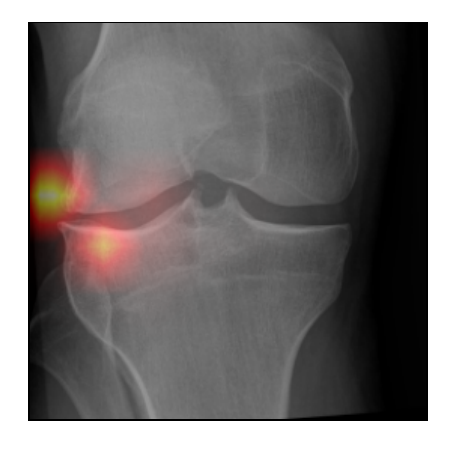}
        \hfill
        \subfloat[]{
        \includegraphics[width=0.26\textwidth]{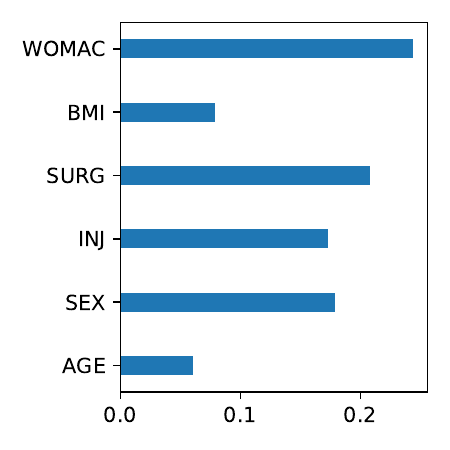}
        } \hfill 
        \includegraphics[width=0.43\textwidth]{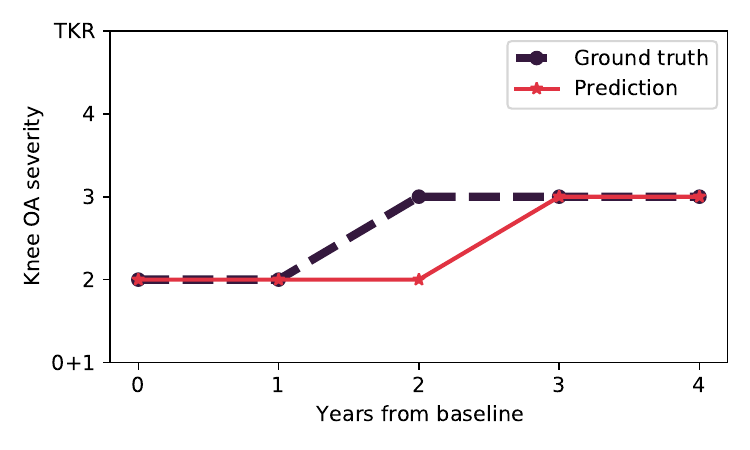}
    \caption{Selective samples of predictions done by CLIMAT from OAI. Picked imaging features corresponded to known clinical findings -- joint space narrowing and osteophytes.  Furthermore, other findings (known in the literature) such as the changes in the intercondylar notch and attrition were also picked by the model.  }
    \label{fig:oai_interpretability_more}
\end{figure*}

\end{document}